\def\year{2021}\relax
\documentclass[letterpaper]{article} % DO NOT CHANGE THIS
\usepackage{aaai21}  % DO NOT CHANGE THIS
\usepackage{times}  % DO NOT CHANGE THIS
\usepackage{helvet} % DO NOT CHANGE THIS
\usepackage{courier}  % DO NOT CHANGE THIS
\usepackage[hyphens]{url}  % DO NOT CHANGE THIS
\usepackage{graphicx} % DO NOT CHANGE THIS
\usepackage{natbib}  % DO NOT CHANGE THIS AND DO NOT ADD ANY OPTIONS TO IT
\usepackage{caption} % DO NOT CHANGE THIS AND DO NOT ADD ANY OPTIONS TO IT
\usepackage{algorithm}
\usepackage{algorithmic}
\usepackage{microtype}
\usepackage{subfig}
\usepackage{booktabs} % for professional tables
\usepackage{amsfonts}       % blackboard math symbols
\usepackage{nicefrac}       % compact symbols for 1/2, etc.
\usepackage{amsmath}
\usepackage{soul}
\usepackage{siunitx}
\usepackage[dvipsnames]{xcolor}

\newcommand{\cut}[1]{}

\title{Improved Consistency Regularization for GANs}

\def\authspace{\hspace{2mm}}

\author {
    Zhengli Zhao\textsuperscript{\rm 1,2}\authspace{}
    Sameer Singh\textsuperscript{\rm 1}\authspace{}
    Honglak Lee\textsuperscript{\rm 2}\authspace{}
    Zizhao Zhang\textsuperscript{\rm 2}\authspace{}
    Augustus Odena\textsuperscript{\rm 2}\authspace{}
    Han Zhang\textsuperscript{\rm 2} \\
}
\affiliations {
    \textsuperscript{\rm 1} University of California, Irvine \\
    \textsuperscript{\rm 2} Google Research \\
    \texttt{zhengliz@uci.edu, zhanghan@google.com}
}

\begin{document}

\maketitle

\begin{abstract}

Recent work \citep{CRGAN} has increased the performance of Generative Adversarial Networks (GANs) by enforcing a 
consistency cost on the discriminator. 
We improve on this technique in several ways. 
We first show that consistency regularization can introduce artifacts into the GAN samples and explain how to fix this issue. 
We then propose several modifications to the consistency regularization procedure designed to improve its performance. 
We carry out extensive experiments quantifying the benefit of our improvements. 
For unconditional image synthesis on CIFAR-10 and CelebA, our modifications yield the best known FID scores on various GAN architectures. 
For conditional image synthesis on CIFAR-10, we improve the state-of-the-art FID score from 11.48 to 9.21. 
Finally, on ImageNet-2012, we apply our technique to the original BigGAN \citep{BigGAN} model and improve the FID from 6.66 to 5.38, which is the best score at that model size.

% Generative Adversarial Networks (GANs) prosper as a powerful class of deep generative models, but are known for the training difficulties.
% Despite the recent advance with consistency regularization~\citep{CRGAN} outperforms other baseline regularizers, it makes the training of the discriminator imbalanced and induces artifacts in generation.
% In this work, we improve the consistency regularization by augmenting both generated and real images passed into the discriminator, and also perturbing the latent vectors fed into the generator.
% We penalize the sensitivity of the discriminator to corresponding image pairs, while encourage the diversity of the generator outputs.
% We carry out systematic experiments to show that our improved consistency regularization (ICR) alleviates generation artifacts and works effectively in different settings.
% For unconditional generation, we achieve the best FID scores on CIFAR-10 and CelebA for various GAN architectures.
% We also improve the state-of-the-art FID scores for conditional generation from 11.48 to 9.21 on CIFAR-10 and from 6.66 to 5.38 on ImageNet-2012.

\end{abstract}

\section{Introduction}

% \zhengli{I'm thinking of framing this work less obvious/straightforward as an extension to CRGAN. Maybe we can first say the motivation/purpose of this work. Then say while there are existing works including CRGAN, which covers a subset of what we are presenting here. We make improvement in several ways and advance performance... Not sure if this can impress the reviewers better, but happy to discuss more.}

% Discriminator regularization has been demonstrated critical to stabilize Generative Adversarial Networks (GANs) \citep{}. Upon many regularization techniques, consistency-based regularization \citep{} recently demonstrates high effectiveness. 
% \augustus{tried to make suggested changes. feel free to add whatever other changes you like.}

Generative Adversarial Networks~\citep[GANs;][]{GAN} are a powerful class of deep generative models, but are known for training difficulties~\citep{salimans2016improved}.
Many approaches have been introduced to improve GAN performance~\citep{wgan, gulrajani2017improved, SNGAN, TOPKGAN}.
Recent work \citep{improvingimproving,CRGAN} suggests that the performance
of generative models can be improved by introducing consistency regularization 
techniques -- which are popular in the semi-supervised learning literature \citep{realisticssl}.
In particular, \citet{CRGAN} show that Generative Adversarial 
Networks (GANs) \citep{GAN} augmented with consistency regularization can achieve 
state-of-the-art image-synthesis results.
In CR-GAN, real images and their corresponding augmented counterparts are fed into the discriminator.
The discriminator is then encouraged --- via an auxiliary loss term --- 
to produce similar outputs for an image and its corresponding augmentation.

Though the consistency regularization in CR-GAN is effective, the augmentations are only applied to the real images
and not to generated samples, making the whole procedure somewhat imbalanced.
In particular, the generator can learn these artificial augmentation features and introduce them into generated samples as undesirable artifacts.\footnote{We show examples in Fig.~\ref{fig:artifact_analysis} and discuss further in Section~\ref{sec:discussion:artifact}.}
Further, by regularizing only the discriminator, and by only using augmentations in image space, 
the regularizations in \citet{improvingimproving} and \citet{CRGAN} do not act directly on the generator.
By constraining the mapping from the prior to the generated samples, we can achieve further performance gains 
on top of those yielded by performing consistency regularization on the discriminator in the first place.

In this work, we introduce \emph{Improved Consistency Regularization}~(ICR) which applies forms 
of consistency regularization to the generated images, the latent vector space, and the generator.
%
% \begin{figure}
%   \centering
%   \subfloat[CIFAR-10 data with $8 \times 8$ cutout.]
%   {\includegraphics[width=0.3\columnwidth]{figures/cutout_aug_8.png}}  \hfill
%   \subfloat[Samples of imbalanced CR-GAN.]
%   {\includegraphics[width=0.3\columnwidth]{figures/cutout_CR_8.png}}  \hfill  
%   \subfloat[Samples of our bCR (Algorithm~\ref{alg:bCR}).]
%   {\includegraphics[width=0.3\columnwidth]{figures/cutout_bCR_8.png}}
% \caption{Illustration of resolving generation artifacts when CIFAR-10 images are augmented with cutout as in (a).
% (b) CR-GAN \citep{CRGAN} can cause augmentation artifacts to appear in generated samples
% with imbalance consistency regularization only on real image augmentations.
% (c) With our Balanced Consistency Regularization (bCR in Algorithm~\ref{alg:bCR}), this issue is fixed with
% both real and generated fake images augmented before fed into the discriminator. 
% }
% \label{fig:artifact_illustration}  
% \end{figure}
% 
% 
First, we address the lack of regularization on the generated samples by introducing 
\emph{balanced consistency regularization} (bCR),
where a consistency term on the discriminator is applied to both real images and samples coming from the generator. 
Second, we introduce \emph{latent consistency regularization} (zCR), which incorporates regularization terms modulating
the sensitivity of both the generator and discriminator changes in the prior.
In particular, given augmented/perturbed latent vectors, we show that it is helpful to encourage the generator to be \emph{sensitive} to the perturbations and the discriminator to be \emph{insensitive}.
We combine bCR and zCR, and call it Improved Consistency Regularization (ICR). 

ICR yields state-of-the-art image synthesis results.
For unconditional image synthesis on CIFAR-10 and CelebA, our method yields the best known FID scores on various GAN architectures. 
For conditional image synthesis on CIFAR-10, we improve the state-of-the-art FID score from 11.48 to 9.21. 
Finally, on ImageNet-2012, we apply our technique to the original BigGAN~\citep{BigGAN} model and improve the 
FID from 6.66 to 5.38, which is the best score at that model size.

\cut{
In summary, the contributions of this work are:

\begin{itemize}
\item We show that the vanilla CR-GAN~\citep{CRGAN} results in augmentation artifacts
in generated samples.
To address this issue, we propose Balanced Consistency 
Regularization (bCR), in which consistency 
regularization is also applied to samples from the generator.

\item In order to more thoroughly adapt consistency regularization techniques to the 
GAN setting, we propose Latent Consistency Regularization (zCR), 
extending the regularization from the discriminator to the generator and from images to latent space.
With draws from the latent prior $z$ perturbed, 
the generator is encouraged to be sensitive to those perturbations, while the discriminator is
encouraged to be consistent.

\item Through a combination of these modifications (which we call Improved Consistency Regularization (ICR)),
we show how GANs can be made to achieve state-of-the-art FID results.
For unconditional image synthesis on CIFAR-10 and CelebA, our method yields the best known FID scores on various GAN architectures. 
For conditional image synthesis on CIFAR-10, we improve the state-of-the-art FID score from 11.48 to 9.21. 
Finally, on ImageNet-2012, we apply our technique to the original BigGAN~\citep{BigGAN} model and improve the FID from 6.66 to 5.38, which is the best score at that model size.
\end{itemize}
}

\begin{figure*}[tb]
\centering
\includegraphics[clip,trim=20 20 20 20,width=0.95\linewidth]{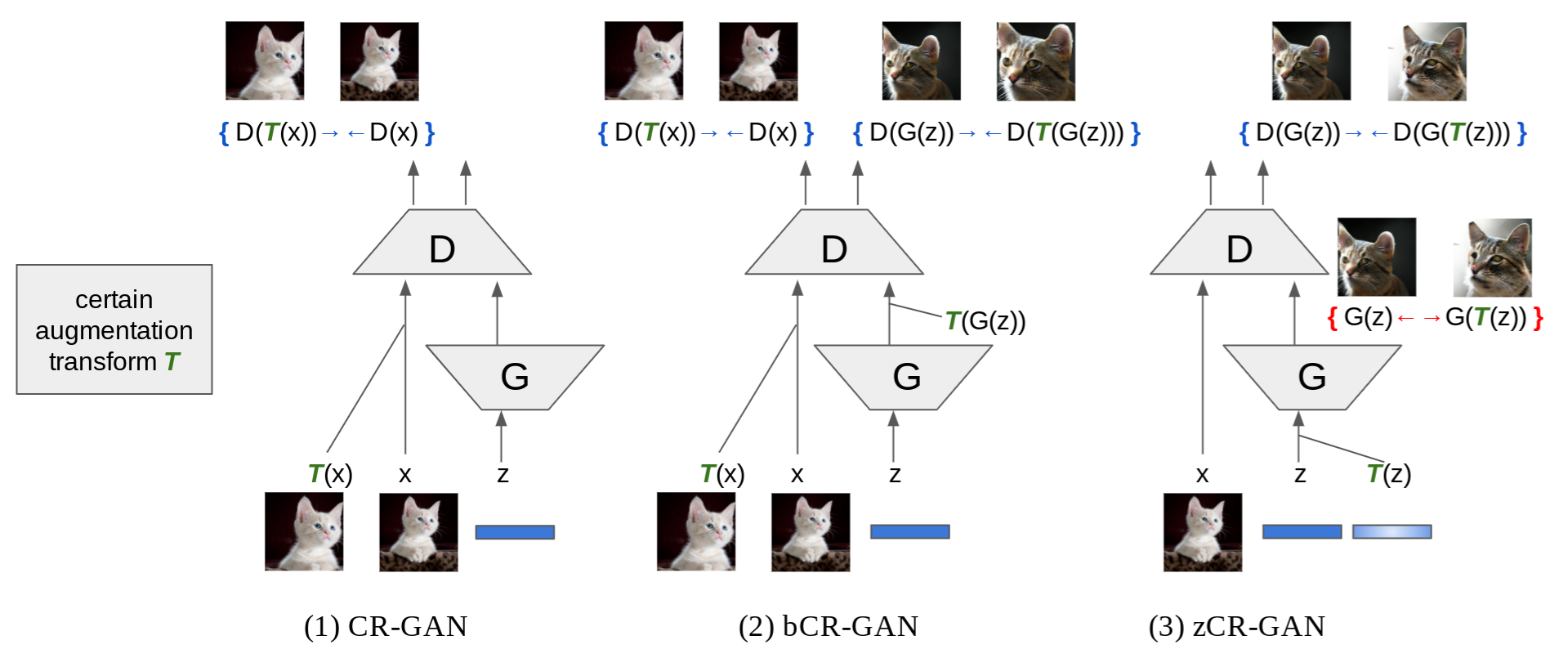}
\caption{\textbf{Illustrations comparing our methods to the baseline.} 
(1) CR-GAN~\citep{CRGAN} is the baseline, with consistency regularization applied only between real images and their augmentations. 
(2) In Balanced Consistency Regularization (bCR-GAN), we also introduce consistency regularization between generated fake images and their augmentations. With consistency regularization on both real and fake images, the discriminator is trained in a balanced way and less augmentation artifacts are generated. 
(3) Furthermore, we propose Latent Consistency Regularization (zCR-GAN), where latent $z$ is augmented with noise of small magnitude. Then for the discriminator, we regularize the consistency between corresponding pairs; while for the generator we encourage the corresponding generated images to be more diverse.
Note that
\textcolor{blue}{\{$\rightarrow\leftarrow$\}} indicates a loss term encouraging pairs to be closer together,
while \textcolor{red}{\{$\leftarrow\rightarrow$\}} indicates a loss term pushing pairs apart.
}
\label{fig:illustration}
% \end{center}
\end{figure*}

\section{Improved Consistency Regularization}

For semi-supervised or unsupervised learning, consistency regularization techniques are effective and have become broadly used recently~\citep{CONSISTENCY, laine2016temporal, CONSISTENCYAVITAL, UDACONSISTENCY, mixmatch2019}.
The intuition behind these techniques is to encode into model training some prior knowledge: 
that the model should produce consistent predictions given input instances and their semantics-preserving augmentations.
The augmentations (or transformations) can take many forms, such as image flipping and rotating, sentence back-translating, or even adversarial attacks.
Penalizing the inconsistency can be easily achieved by minimizing $L_2$ loss \citep{CONSISTENCY,laine2016temporal} between instance pairs, or KL-divergence loss \citep{UDACONSISTENCY, miyato2018virtual} between distributions.
In the GAN literature, \citet{improvingimproving} propose a consistency term derived from Lipschitz 
continuity considerations to improve the training of WGAN.
Recently, CR-GAN~\citep{CRGAN} applies consistency regularization to the discriminator and achieves substantial improvements.

Below we start by introducing our two new techniques, abbreviated as bCR and zCR, to improve and generalize CR for GANs. 
We denote the combination of both of these techniques as ICR, and we will later show that ICR
yields state-of-the-art image synthesis results in a variety of settings.
Figure~\ref{fig:illustration} shows illustrations comparing our methods to the baseline CR-GAN~\citet{CRGAN}.

\subsection{Balanced Consistency Regularization (bCR)}
\label{sec:bCR}

\begin{algorithm}[tb]
\small
\caption{Balanced Consistency Regularization (bCR)}
\label{alg:bCR}
\begin{algorithmic}
    \STATE {\bfseries Input:} parameters of generator $\theta_G$ and discriminator $\theta_D$, consistency regularization coefficient for real images $\lambda_\text{real}$ and fake images $\lambda_\text{fake}$, 
    % number of discriminator iterations per generator iteration $N_D$, 
    augmentation transform $T$ (for images, e.g. shift, flip, cutout, etc).
    \FOR{number of training iterations}
    % \FOR{$t=1$ {\bfseries to} $N_D$}
    \STATE Sample batch $z \sim p(z)$, $x \sim p_{\text{real}}(x)$
    \STATE Augment both real $T(x)$ and fake $T(G(z))$ images
    \STATE $L_D \leftarrow D(G(z)) - D(x)$
    \STATE $L_\text{real} \leftarrow \lVert D(x) - D(T(x)) \rVert ^2$
    \STATE $L_\text{fake} \leftarrow \lVert D(G(z)) - D(T(G(z))) \rVert ^2$
    \STATE $\theta_D \leftarrow \text{AdamOptimizer}(L_D + \lambda_{\text{real}} L_\text{real} + \lambda_{\text{fake}} L_\text{fake})$
    % \ENDFOR
    % \STATE Sample batch $z \sim p(z)$
    \STATE $L_G \leftarrow -D(G(z))$
    \STATE $\theta_G \leftarrow \text{AdamOptimizer}(L_G)$
    \ENDFOR
\end{algorithmic}
\end{algorithm}

Figure~\ref{fig:illustration}(1) illustrates the baseline CR-GAN,
in which a term is added to the discriminator loss function that penalizes its sensitivity to the difference between the original image $x$ and the augmented image $T(x)$.
One key problem with the original CR-GAN is that the discriminator might `mistakenly believe' that the augmentations are
actual features of the target data set, since these augmentations are only performed on the real images.
This phenomenon, which we refer to as consistency imbalance, is not easy to notice for certain types of 
augmentation (e.g. image shifting and flipping). 
However, it can result in generated samples with explicit augmentation artifacts when augmented samples contain visual artifacts not belonging to real images. 
For example, we can easily observe this effect for CR-GAN with cutout augmentation: see the second column in Figure~\ref{fig:artifact_analysis}. 
This undesirable effect greatly limits the choice of advanced augmentations we could use.

In order to correct this issue, we propose to also augment generated samples before they are fed into the discriminator,
so that the discriminator will be evenly regularized with respect to both real and fake augmentations and thereby 
be encouraged to focus on meaningful visual information.  

Specifically, a gradient update step will involve four batches, a batch of real images $x$, 
augmentations of these real images $T(x)$, a batch of generated samples $G(z)$, 
and that same batch with augmentations $T(G(z))$.
The discriminator will have terms that penalize
its sensitivity between corresponding $\{x, T(x)\}$ and also $\{G(z), T(G(z))\}$, 
while the generator cost remains unmodified.

This technique is described in more detail in Algorithm \ref{alg:bCR} and visualized in Figure~\ref{fig:illustration}(2).
We abuse the notation a little in the sense that $D(x)$ denotes the output vector before 
activation of the last layer of the discriminator given input $z$.
$T(x)$ denotes an augmentation transform, here for images (e.g. shift, flip, cutout, etc).
The consistency regularization can be balanced by adjusting the 
strength of $\lambda_{\text{real}}$ and $\lambda_{\text{fake}}$.
This proposed bCR technique not only removes augmentation artifacts (see third column of
Figure~\ref{fig:artifact_analysis}), but also brings substantial performance improvement 
(see Section~\ref{sec:experiments}~and~\ref{sec:discussion}).

\subsection{Latent Consistency Regularization (zCR)}

\begin{algorithm}[hb]
\small
\caption{Latent Consistency Regularization (zCR)}
\label{alg:zCR}
\begin{algorithmic}
    \STATE {\bfseries Input:} parameters of generator $\theta_G$ and discriminator $\theta_D$, consistency regularization coefficient for generator $\lambda_\text{gen}$ and discriminator $\lambda_\text{dis}$, 
    % number of discriminator iterations per generator iteration $N_D$, 
    augmentation transform $T$ (for latent vectors, e.g. adding small perturbation noise$\sim \mathcal{N}(0, \sigma_\text{noise})$).
    \FOR{number of training iterations}
    % \FOR{$t=1$ {\bfseries to} $N_D$}
    \STATE Sample batch $z \sim p(z)$, $x \sim p_{\text{real}}(x)$
    \STATE Sample perturbation noise $\Delta z \sim \mathcal{N}(0, \sigma_\text{noise})$
    \STATE Augment latent vectors $T(z) \leftarrow z + \Delta z$ 
    \STATE $L_D \leftarrow D(G(z)) - D(x)$
    \STATE $L_\text{dis} \leftarrow \lVert D(G(z)) - D(G(T(z))) \rVert ^2$
    \STATE $\theta_D \leftarrow \text{AdamOptimizer}(L_D + \lambda_{\text{dis}} L_\text{dis})$
    % \ENDFOR
    % \STATE Sample batch $z \sim p(z)$
    \STATE $L_G \leftarrow -D(G(z))$
    \STATE $L_\text{gen} \leftarrow -\lVert G(z) - G(T(z)) \rVert ^2$
    \STATE $\theta_G \leftarrow \text{AdamOptimizer}(L_G + \lambda_{\text{gen}} L_\text{gen})$
    \ENDFOR
\end{algorithmic}
\end{algorithm}

In Section~\ref{sec:bCR}, we focus on consistency regularization with respect to augmentations in image space
on the inputs to the discriminator. 
In this section, we consider a different question:
Would it help if we enforce consistency regularization on augmentations in latent space~\citep{GNAE}?
Given that a GAN model consists of both a generator and a discriminator, it seems reasonable to ask
if techniques that can be applied to the discriminator can also be effectively applied to the generator in certain analogous way.

Towards this end, we propose to augment inputs to the generator 
by slightly perturbing draws $z$ from the prior to yield 
$T(z) = z + \Delta z, \Delta z \sim \mathcal{N}(0, \sigma_\text{noise})$.
Assuming the perturbations $\Delta z$ are small enough, 
we expect that output of the discriminator ought not to change much with respect to this perturbation
and modify the discriminator loss by enforcing $\lVert D(G(z)) - D(G(T(z))) \rVert ^2$ is small.

However, with only this term added onto the GAN loss,
the generator would be prone to collapse to generating specific samples for any latent $z$,
since that would easily satisfy the constraint above. 
To avoid this, we also modify the loss function for the generator 
with a term that maximizes the difference between $G(z)$ and $G(T(z))$,
which also encourages generations from similar latent vectors to be diverse.
Though motivated differently, this can be seen as related to the Jacobian Clamping technique from \citet{oboborg} and diversity increase technique in \citet{yang2019diversity}.

This method is described in more detail in Algorithm \ref{alg:zCR} and visualized in Figure~\ref{fig:illustration}(3).
$G(z)$ denotes the output images of the generator given input $z$.
$T(x)$ denotes an augmentation transform, here for latent vectors (e.g. adding small perturbation noise).
The strength of consistency regularization for the discriminator can be adjusted via $\lambda_{\text{dis}}$.
From the view of the generator, intuitively, 
the term $L_\text{gen} = -\lVert G(z) - G(T(z)) \rVert ^2$ encourages $\{G(z), G(T(z))\}$ to be diverse.
We have conducted analysis on the effect of $\lambda_\text{gen}$ with experiments in Section~\ref{sec:discussion:zCR}.
This technique substantially improves the performance of GANs, as measured by FID.
We present experimental results in Section~\ref{sec:experiments}~and~\ref{sec:discussion}.

\subsection{Putting it All Together (ICR)}
Though both Balanced Consistency Regularization and Latent Consistency Regularization improve GAN performance 
(see Section \ref{sec:experiments}), it is not obvious that they would work when `stacked on top' of each other.
That is, maybe they are accomplishing the same thing in different ways, and we cannot add up their benefits.
However, validated with extensive experiments,
we achieve the best experimental results when combining Algorithm~\ref{alg:bCR} 
and Algorithm~\ref{alg:zCR} together.
We call this combination Improved Consistency Regularization (ICR).
Note that in ICR, we augment inputs in both image and latent spaces, 
and add regularization terms to both the discriminator and the generator.
We regularize the discriminator's consistency between corresponding pairs of
$\{D(x), D(T(x))\}$, $\{D(G(z)), D(T(G(z)))\}$, and $\{D(G(z)), D(G(T(z)))\}$;
For the generator, we encourage diversity between $\{G(z), G(T(z))\}$.

\section{Experiments}
\label{sec:experiments}

In this section, we validate our methods on different data sets,
model architectures, and GAN loss functions.
We compare both Balanced Consistency Regularization (Algorithm~\ref{alg:bCR}) and Latent Consistency Regularization (Algorithm~\ref{alg:zCR}) with several baseline methods.
We also combine both techniques (we abbreviate this combination as ICR) and show
that this yields state-of-the-art FID numbers.
We follow the best experimental practices established in \citet{compare_gan},
aggregating all runs and reporting the FID distribution of the top 15\% of trained models.
We provide both quantitative and qualitative results (with more in the appendix).

\subsection{Baseline Methods}

We compare our methods with four GAN regularization techniques: 
Gradient Penalty (GP)~\citep{gulrajani2017improved},
DRAGAN (DR)~\citep{kodali2017convergence},
Jensen-Shannon Regularizer (JSR)~\citep{roth2017stabilizing},
and vanilla Consistency Regularization (CR)~\citep{CRGAN}.
The regularization strength $\lambda$ is set to 0.1 for JSR, and 10 for all others.

Following the procedures from \citet{lucic2018gans, compare_gan},
we evaluate these methods across different data sets,
neural architectures, and loss functions. 
For optimization, we use the Adam optimizer with batch size of 64 for all experiments.
By default, spectral normalization (SN)~\citep{SNGAN} is used in the
discriminator, as it is the most effective normalization method for 
GANs~\citep{compare_gan} and is becoming the 
standard for recent GANs~\citep{BigGAN, LOGAN}.

\subsection{Data Sets and Evaluation}

We carry out extensive experiments comparing our methods against the above 
baselines on three commonly used data sets in the GAN literature:
CIFAR-10~\citep{CIFAR},
CelebA-HQ-128~\citep{PGGAN},
and ImageNet-2012~\citep{ImageNet}.

For data set preparation, we follow the detailed procedures in \citet{compare_gan}.
CIFAR-10 contains 60K $32 \times 32$ images with 10 labels, 
out of which 50K are used for training and 10K are used for testing.
CelebA-HQ-128 (CelebA) consists of 30K $128 \times 128$ facial images, 
out of which we use 3K images for testing and train models with the rest.
ImageNet-2012 has approximately 1.2M images with 1000 labels, 
and we down-sample the images to $128 \times 128$.
We stop training after 200k generator update steps for CIFAR-10,
100k steps for CelebA, and 250k for ImageNet.

We use the Fréchet Inception Distance (FID)~\citep{FID} as 
the primary metric for quantitative evaluation.
FID has been shown to correlate well with human evaluation of image quality
and to be helpful in detecting intra-class mode collapse.
We calculate FID between generated samples and real test images,
using 10K images on CIFAR-10, 3K on CelebA, and 50K on ImageNet.
We also report Inception Scores~\citep{salimans2016improved} in the appendix.

By default, the augmentation transform $T$ on latent vectors $z$ is adding Gaussian 
noise $\Delta z \sim \mathcal{N}(0, \sigma_\text{noise})$.
The augmentation transform $T$ on images is a combination of randomly flipping horizontally and shifting by multiple pixels (up to 4 for CIFAR-10 and 
CelebA, and up to 16 for ImageNet). 
This transform combination results in better performance than alternatives (see \citet{CRGAN}).
Though we outperform CRGAN for different augmentation strategies, we use the same image augmentation strategies as the best one (random flip and shift) in CRGAN for comparison.

There are many different GAN loss functions and we elaborate on several of them in Section~\ref{sec:background}.
Following \citet{CRGAN}, for each data set and model architecture combination,
we conduct experiments using the loss function that achieves the best performance
on baselines.

\subsection{Unconditional GAN Models}
\label{sec:unconditional}

We first test out techniques on unconditional image
generation,
which is to model images from an 
object-recognition data set without any reference to the underlying
classes.
We conduct experiments on the CIFAR-10 and CelebA data sets,
and use both DCGAN~\citep{DCGAN} and ResNet~\citep{resnet} GAN architectures.

\subsubsection{DCGAN on CIFAR-10}

\begin{figure}[tb]
\begin{center}
\centerline{\includegraphics[width=\columnwidth]{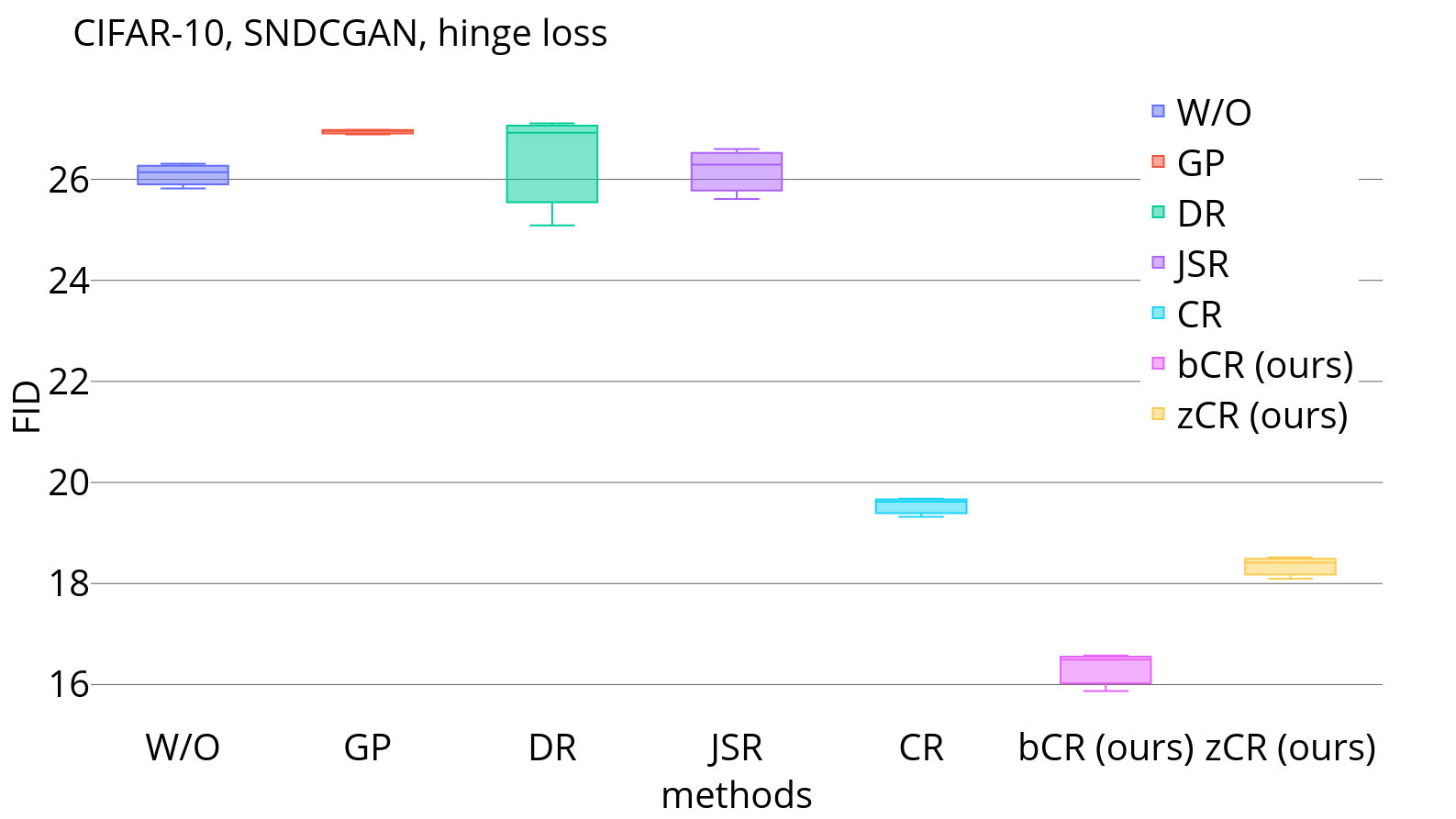}}
\caption{
\textbf{FID scores for DCGAN trained on CIFAR-10} with the
hinge loss, for a variety of regularization techniques.
Consistency regularization significantly outperforms non-consistency regularizations.
Adding Balanced Consistency Regularization causes a larger improvement than 
Latent Consistency Regularization, but both yield improvements much larger 
than measurement variances.
}
\label{fig:cifar_dcgan_hinge}
\end{center}
\end{figure}

Figure~\ref{fig:cifar_dcgan_hinge} presents the results of DCGAN on CIFAR-10 
with the hinge loss.
Vanilla Consistency Regularization (CR) \citep{CRGAN} outperforms all other baselines.
Our Balanced Consistency Regularization (bCR) technique 
improves on CR by more than $3.0$ FID points.
Our Latent Consistency Regularization (zCR) technique 
improves scores less than bCR,
but the improvement is still significant compared to the measurement variance.
We set $\lambda_\text{real} = \lambda_\text{fake} = 10$ for bCR,
while using $\sigma_\text{noise} = 0.03$, $\lambda_\text{gen} = 0.5$,
and $\lambda_\text{dis} = 5$ for zCR.

\begin{figure}[tb]
\begin{center}
\centerline{\includegraphics[width=\columnwidth]{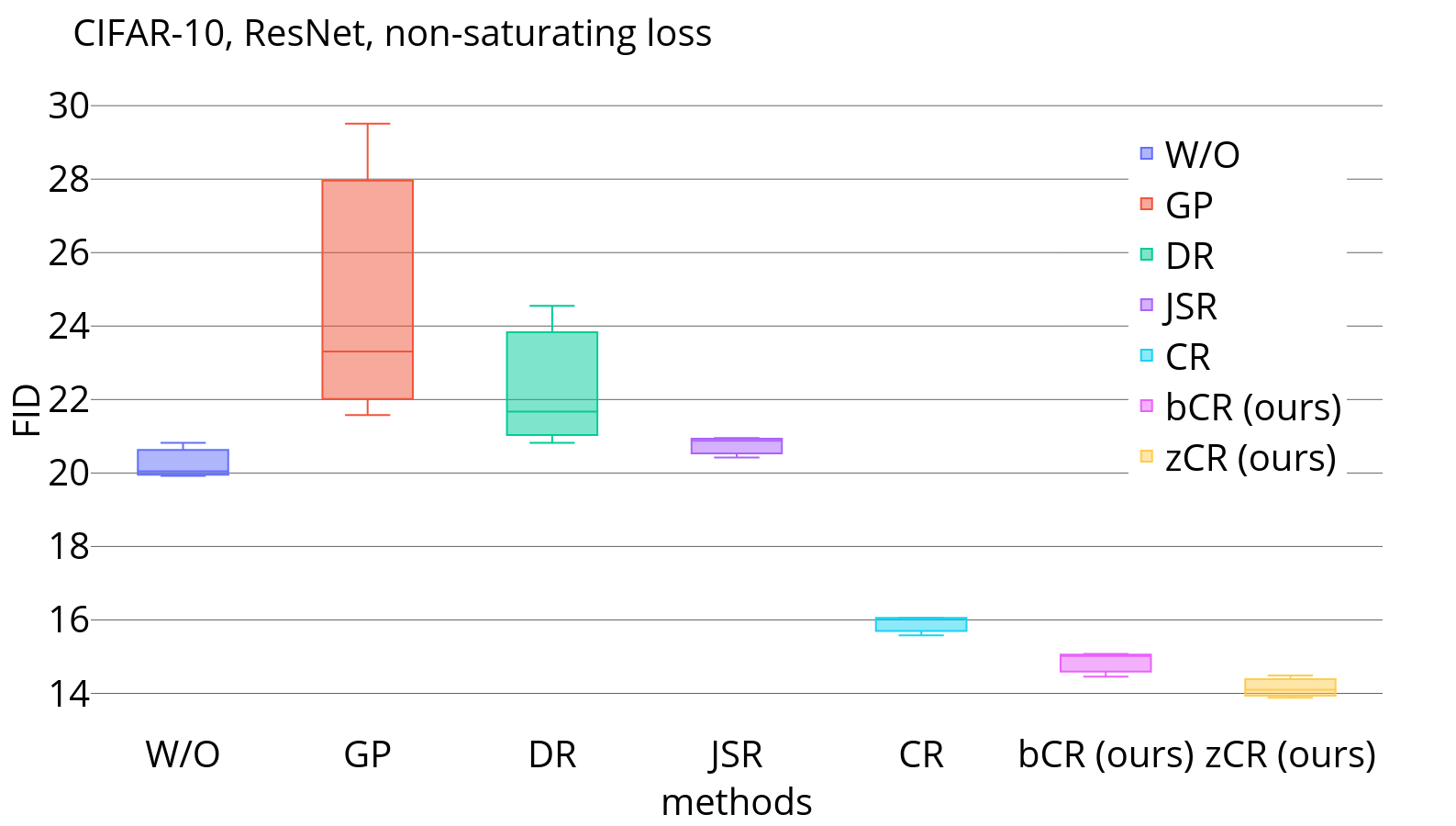}}
\caption{
\textbf{FID scores for a ResNet-style GAN trained on CIFAR-10} with the non-saturating loss, for a variety of regularization techniques.
Contrary to the results in Figure~\ref{fig:cifar_dcgan_hinge},
Latent Consistency Regularization outperforms Balanced Consistency Regularization,
though they both substantially surpass all baselines.
}
\label{fig:cifar_resnet_ns}
\end{center}
\end{figure}

\subsubsection{ResNet on CIFAR-10}

DCGAN-type models are well-known and it is 
encouraging that our techniques increase performance for those models, 
but they have been substantially surpassed in performance by newer techniques.
We then validate our methods on more recent architectures that use residual 
connections \citep{resnet}.
Figure~\ref{fig:cifar_resnet_ns} shows unconditional image synthesis results on 
CIFAR-10 using a GAN model with residual connections and the non-saturating loss.
Though both of our proposed modifications still outperform all baselines,
Latent Consistency Regularization works better than Balanced Consistency Regularization,
contrary to the results in Figure \ref{fig:cifar_dcgan_hinge}.
For hyper-parameters, 
we set $\lambda_\text{real} = 10$ and $\lambda_\text{fake} = 5$ for bCR,
while using $\sigma_\text{noise} = 0.07$, $\lambda_\text{gen} = 0.5$,
and $\lambda_\text{dis} = 20$ for zCR.

\subsubsection{DCGAN on CelebA}

\begin{figure}[t]
\begin{center}
\centerline{\includegraphics[width=\columnwidth]{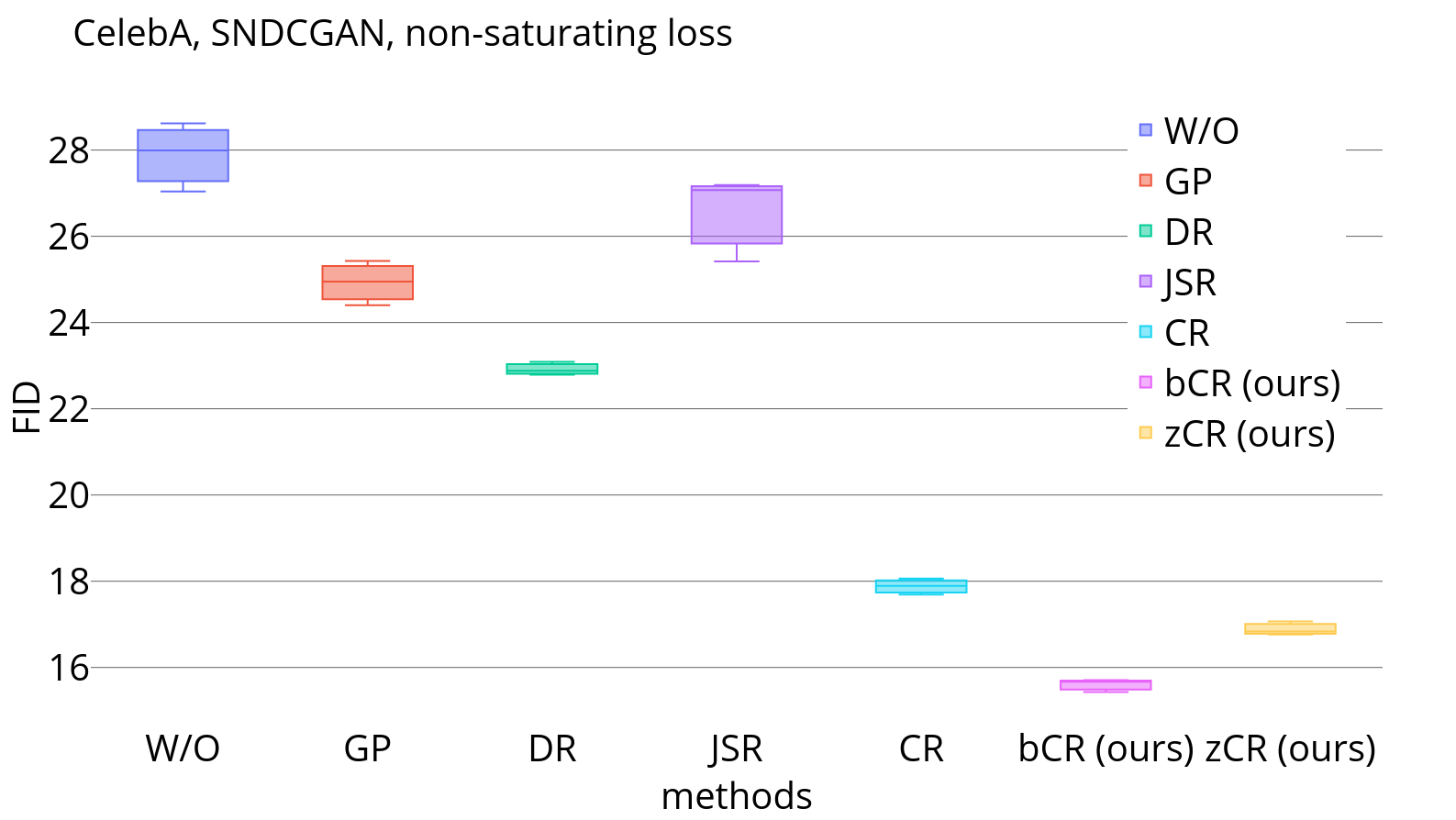}}
\caption{
\textbf{FID scores for DCGAN trained on CelebA} with the non-saturating loss, for a variety of regularization techniques.
Consistency regularization significantly outperforms all other baselines.
Balanced Consistency Regularization further improves on Consistency Regularization by more than 2.0 in terms of FID,
while Latent Consistency Regularization improves by around 1.0.
}
\label{fig:celeba_dcgan_ns}
\end{center}
\end{figure}

We also conduct experiments on the CelebA data set.
The baseline model we use in this case is a DCGAN model with the non-saturating loss.
We set $\lambda_\text{real} = \lambda_\text{fake} = 10$ for bCR,
while using $\sigma_\text{noise} = 0.1$, $\lambda_\text{gen} = 1$, 
and $\lambda_\text{dis} = 10$ for zCR.
The results are shown in Figure~\ref{fig:celeba_dcgan_ns} and are overall similar
to those in Figure~\ref{fig:cifar_dcgan_hinge}.
The improvements in performance for CelebA are not as large as those for 
CIFAR-10, 
but they are still substantial,
suggesting that our methods generalize across data sets.

% \paragraph{Combining Balanced and Latent Consistency Regularization}
\subsubsection{Improved Consistency Regularization}
As alluded to above, 
we observe experimentally that 
combining Balanced Consistency regularization (bCR) and Latent Consistency 
Regularization (zCR) (into Improved Consistency Regularization (ICR))
yields results that are better than those given by either method alone.
Using the above experimental results,
we choose the best-performing hyper-parameters to carry out experiments for ICR, regularizing with both bCR and zCR.
Table~\ref{tab:unconditional} shows 
that ICR yields the best results for all three unconditional synthesis settings we study.
Moreover, the results of the ResNet model on CIFAR-10 are, to the best of our knowledge,
the best reported results for unconditional CIFAR-10 synthesis.

\begin{table}[tb]
\caption{\textbf{FID scores for Unconditional Image Synthesis.}
ICR achieves the best performance overall.
Baselines are: not using regularization (W/O),
Gradient Penalty (GP)~\citep{gulrajani2017improved},
DRAGAN (DR)~\citep{kodali2017convergence},
Jensen-Shannon Regularizer (JSR)~\citep{roth2017stabilizing},
and vanilla Consistency Regularization (CR)~\citep{CRGAN}.
}
\label{tab:unconditional}
\begin{center}
\begin{tabular}{lccc}
\toprule
        & CIFAR-10  & CIFAR-10  & CelebA \\
Methods & (DCGAN) & (ResNet)  & (DCGAN) \\
\midrule
W/O     & 24.73 & 19.00 & 25.95 \\
GP      & 25.83 & 19.74 & 22.57 \\
DR      & 25.08 & 18.94 & 21.91 \\
JSR     & 25.17 & 19.59 & 22.17 \\
CR      & 18.72 & 14.56 & 16.97 \\
\midrule
% bCR     & \textbf{15.87} & 14.45 & \textbf{15.43} \\
% zCR     & 18.10 & \textbf{13.36} & 16.76 \\
ICR (ours)    & \textbf{15.87} & \textbf{13.36} & \textbf{15.43} \\
\bottomrule
\end{tabular}
\end{center}
\end{table}

\subsection{Conditional GAN Models}
\label{sec:conditional}

\begin{table}[ht!]
\caption{\textbf{FID scores for class conditional image generation on CIFAR-10 and ImageNet.}
We compare our ICR technique with state-of-the-art GAN models including 
SNGAN~\citep{SNGAN}, BigGAN~\citep{BigGAN}, and CR-GAN~\citep{CRGAN}. 
The BigGAN implementation we use is from \citet{compare_gan}. 
($\ast$)-BigGAN has the exactly same architecture as the publicly available BigGAN and
is trained with the same settings, but with our consistency regularization techniques
added to GAN losses.
On CIFAR-10 and ImageNet, we improve the FID numbers to 9.21 and 5.38 correspondingly, 
which are the best known scores at that model size.}
\label{tab:conditional}
\begin{center}
\begin{tabular}{lcc}
\toprule
Models      & CIFAR-10 & ImageNet \\
\midrule
SNGAN       & 17.50 & 27.62 \\
BigGAN      & 14.73 & 8.73 \\
CR-BigGAN   & 11.48 & 6.66 \\
\midrule
bCR-BigGAN  & 10.54 & 6.24 \\
zCR-BigGAN  & 10.19 & 5.87 \\
ICR-BigGAN  & \textbf{9.21} & \textbf{5.38} \\
\bottomrule
\end{tabular}
\end{center}
\end{table}

In this section, we apply our consistency regularization techniques to the publicly 
available implementation of BigGAN \citep{BigGAN} from \citet{compare_gan}.
We compare it to baselines from \citet{BigGAN,SNGAN,CRGAN}.
Note that the FID numbers from \citet{LOGAN} are based on a larger version of BigGAN 
called BigGAN-Deep with substantially more parameters than the original BigGAN, 
and are thus not comparable to the numbers we report here.

On CIFAR-10, our techniques yield the best known FID score for conditional synthesis with 
CIFAR-10\footnote{
There are a few papers that report lower scores using the PyTorch implementation
of the FID. 
That implementation outputs numbers that are much lower,
which are not comparable to numbers from the official
TF implementation,  
as explained at 
\url{https://github.com/ajbrock/BigGAN-PyTorch\#an-important-note-on-inception-metrics}}: 9.21.
On conditional Image Synthesis on the ImageNet data set, 
our technique yields FID of 5.38.
This is the best known score using the same number of parameters as in the original BigGAN model,
though the much larger model from \citet{LOGAN} achieves a better score.
For both setups, we set $\lambda_\text{real} = \lambda_\text{fake} = 10$, 
together with $\sigma_\text{noise} = 0.05$, $\lambda_\text{gen} = 0.5$, 
and $\lambda_\text{dis} = 20$.

\section{Ablation Studies}
\label{sec:discussion}
To better understand how the various hyper-parameters introduced by 
our new techniques affect performance, we conduct a series of ablation studies.
We include both quantitative and qualitative results.

\subsection{Examining Artifacts Resulting from `Vanilla' Consistency Regularization}
\label{sec:discussion:artifact}

\begin{figure}[bt]
  \centering
  \subfloat[$8 \times 8$ cutout.]
  {\includegraphics[width=0.3\columnwidth]{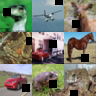}}  \hfill
  \subfloat[CR samples.]
  {\includegraphics[width=0.3\columnwidth]{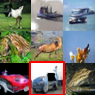}}  \hfill  
  \subfloat[bCR samples.]
  {\includegraphics[width=0.3\columnwidth]{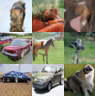}} \hfill  
  \subfloat[$16 \times 16$ cutout.]
  {\includegraphics[width=0.3\columnwidth]{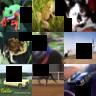}}  \hfill
  \subfloat[CR samples.]
  {\includegraphics[width=0.3\columnwidth]{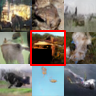}}  \hfill  
  \subfloat[bCR samples.]
  {\includegraphics[width=0.3\columnwidth]{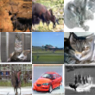}} \hfill
  \subfloat[$32 \times 32$ cutout.]
  {\includegraphics[width=0.3\columnwidth]{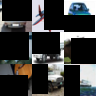}}  \hfill
  \subfloat[CR samples.]
  {\includegraphics[width=0.3\columnwidth]{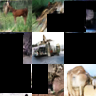}}  \hfill  
  \subfloat[bCR samples.]
  {\includegraphics[width=0.3\columnwidth]{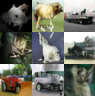}}  
\caption{
\textbf{Illustration of resolving generation artifacts by Balanced Consistency Regularization.}
The first column shows CIFAR-10 training images augmented with cutout of different sizes.
The second column demonstrates that the vanilla CR-GAN~\citep{CRGAN} can cause augmentation artifacts to appear in generated samples.
This is because CR-GAN only has consistency regularization on real images passed into the discriminator.
In the last column (our Balanced Consistency Regularization: bCR in Algorithm~\ref{alg:bCR}) this issue is fixed with both real and generated fake images augmented before being fed into the discriminator.
}
\label{fig:artifact_analysis}  
\end{figure}

\begin{table}[tb]
\caption{\textbf{Fraction of Artifacts:} bCR alleviates generation artifacts the more it is enforced (higher $\lambda_\text{fake}$).}
\label{tab:artifact_analysis}
\begin{center}
\begin{tabular}{lccc}%{p{1.0cm}p{1.8cm}p{1.8cm}p{1.8cm}}
\toprule
$\mathbf{\lambda_\text{fake}}$      & \bf 8x8 cutout & \bf 16x16 cutout & \bf 32x32 cutout \\
\midrule
    0 & $0.07 \pm 0.03$ & $0.12 \pm 0.02$ & $0.66 \pm 0.03$ \\
    2 & $0.03 \pm 0.02$ & $0.08 \pm 0.01$ & $0.09 \pm 0.02$ \\
    5 & $0.01 \pm 0.01$ & $0.05 \pm 0.01$ & $0.03 \pm 0.01$ \\
    10 & $0.01 \pm 0.01$ & $0.01 \pm 0.01$ & $0.01 \pm 0.01$ \\
\bottomrule
\end{tabular}
\end{center}
\end{table}

To understand the augmentation artifacts resulting from using vanilla CR-GAN~\citep{CRGAN}, and to validate
that Balanced Consistency Regularization removes those artifacts, we carry out a series of qualitative
experiments using varying sizes for the cutout \citep{cutout} augmentation.
We experiment with cutouts of size $8 \times 8$, $16 \times 16$, and $32 \times 32$,
training both vanilla CR-GANs and GANs with Balanced Consistency Regularization.
The results are shown in Figure~\ref{fig:artifact_analysis}.
The first column shows CIFAR-10 training images augmented with cutout of different sizes.
The second column demonstrates that the vanilla CR-GAN~\citep{CRGAN} can cause augmentation artifacts to appear in generated samples.
This is because CR-GAN only has consistency regularization on real images passed into the discriminator.
In the last column (our Balanced Consistency Regularization: bCR in Algorithm~\ref{alg:bCR}) this issue is fixed with both real and generated fake images augmented before being fed into the discriminator.
Broadly speaking, we observe more substantial cutout artifacts (black rectangles) 
in samples from CR-GANs with larger cutout augmentations,
and essentially no such artifacts for GANs trained with 
Balanced Consistency Regularization with $\lambda_\text{fake} \geq \lambda_\text{real}$.

To quantify how much bCR alleviates generation artifacts, 
we vary cutout sizes and the strength of CR for generated images. 
We examine 600 of generated images from 3 random runs, and report the fraction of images that contain artifacts of cutouts in Table~\ref{tab:artifact_analysis}.
The strength of CR for real images is fixed at $\lambda_\text{real}=10$.
We do observe a few artifacts when $0 < \lambda_\text{fake} \ll \lambda_\text{real}$,
but much less than those from the vanilla CR-GAN.
We believe that this phenomenon of introducing augmentation artifacts into generations likely 
holds for other types of augmentation, but it is much more difficult
to confirm for less visible transforms, and sometimes it may not actually be harmful (e.g. flipping of images in most contexts).

\subsection{Effect of Hyper-Parameters on Balanced Consistency Regularization's Performance}

In Balanced Consistency Regularization (Algorithm \ref{alg:bCR}), 
the cost associated
with sensitivity to augmentations of the real images is weighted by $\lambda_\text{real}$
and the cost associated with sensitivity to augmentations of the generated samples is 
weighted by $\lambda_\text{fake}$.
In order to better understand the interplay between these parameters, we train 
a DCGAN-type model with spectral normalization on the CIFAR-10 data set with the hinge loss, 
for many different values of $\lambda_\text{fake}, \lambda_\text{real}$.
The heat map in Figure \ref{fig:real_fake_ratio} in the appendix shows that it never pays to set either
of the parameters to zero: this means that Balanced Consistency Regularization always
outperforms vanilla consistency regularization (the baseline CR-GAN).
Generally speaking, setting $\lambda_\text{real}$ and $\lambda_\text{fake}$ similar in 
magnitude works well.
This is encouraging, since it means that the performance of bCR
is relatively insensitive to hyper-parameters.

\subsection{Effect of Hyper-Parameters on Latent Consistency Regularization's Performance}
\label{sec:discussion:zCR}

Latent Consistency Regularization (Algorithm \ref{alg:zCR}) has three hyper-parameters:
$\sigma_\text{noise}, \lambda_\text{gen}$ and $\lambda_\text{dis}$, which respectively 
govern the magnitude of the perturbation made to the draw from the prior, the weight of 
the sensitivity of the generator to that perturbation, and the weight of the sensitivity of
the discriminator to that perturbation.
From the view of the generator, intuitively, 
the extra loss term added 
$L_\text{gen} = -\lVert G(z) - G(T(z)) \rVert ^2$ encourages 
$G(z)$ and $G(T(z))$ to be far away from each other.

\begin{figure}[tb]
\begin{center}
\centerline{\includegraphics[width=\columnwidth]{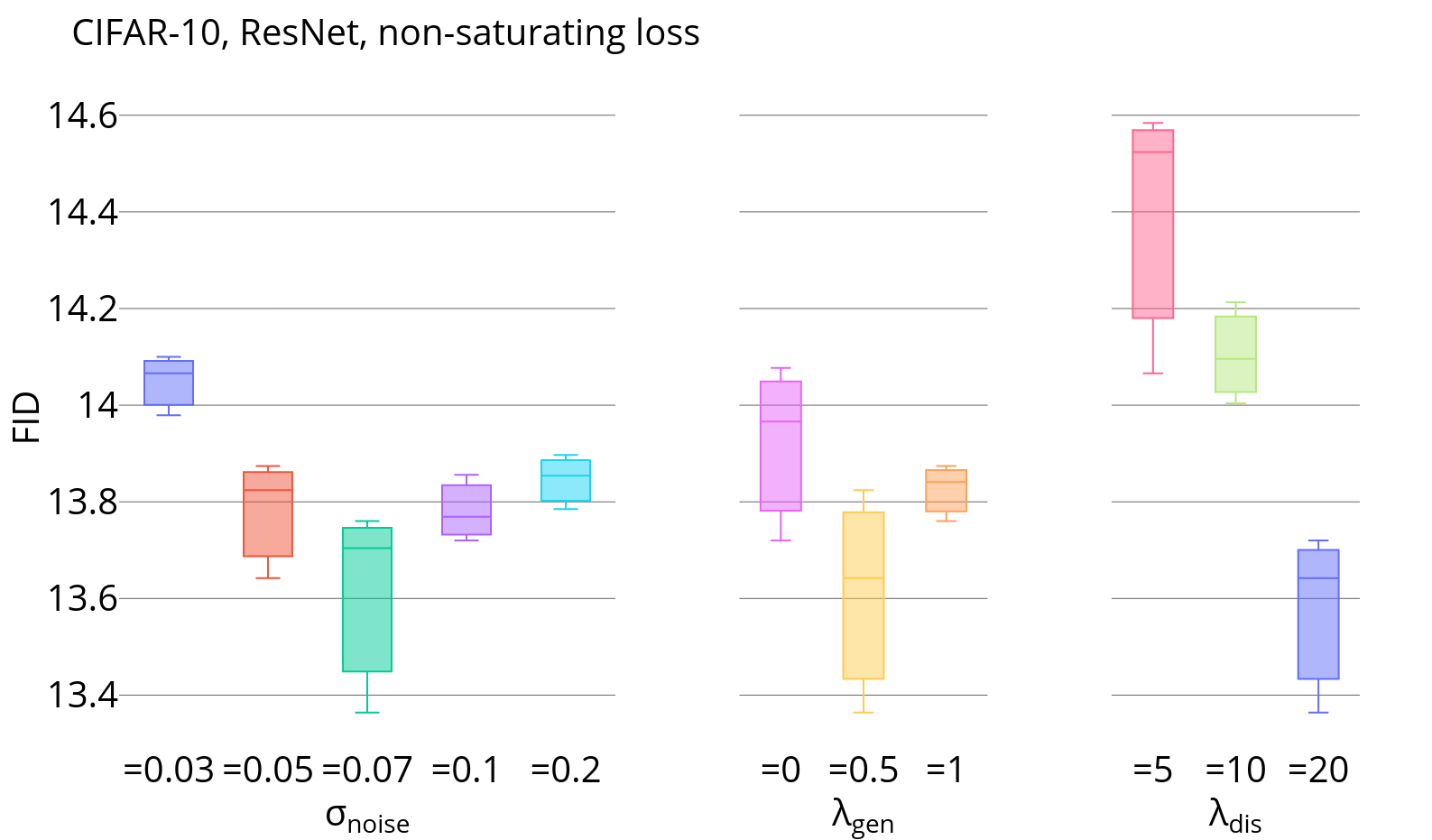}}
\caption{
\textbf{Analysis on the hyper-parameters of Latent Consistency Regularization.}
We conduct experiments using a ResNet-style GAN on CIFAR-10 with  non-saturating loss
in order to better understand the interplay between $\sigma_\text{noise}$, $\lambda_\text{gen}$ and $\lambda_\text{dis}$.
The results show
that a moderate value  for the generator
coefficient (e.g. $\lambda_\text{gen} = 0.5$) works the best.
With the added term $L_\text{gen} = -\lVert G(z) - G(T(z)) \rVert ^2$, 
the generator is encouraged to be sensitive to perturbations in latent space.
For this set of experiments, we observe the best performance adding perturbations with standard deviation of  
$\sigma_\text{noise} = 0.07$, and higher (but not extremely high) values for the discriminator coefficient $\lambda_\text{dis}$ also improve further.
}
\label{fig:z_aug_params}
\end{center}
\end{figure}

We conduct experiments using a ResNet-style GAN on the CIFAR-10 data set with the non-saturating loss
in order to better understand the interplay between these hyper-parameters.
The results in Figure~\ref{fig:z_aug_params} show
that a moderate value for the generator
coefficient (e.g. $\lambda_\text{gen} = 0.5$) works the best (as measured by FID).
This corresponds to encouraging the generator to be sensitive to perturbations
of samples from the prior.
For this experimental setup, perturbations with standard deviation of  
$\sigma_\text{noise} = 0.07$ are the best, and higher (but not extremely high) values for the discriminator coefficient $\lambda_\text{dis}$ also perform better.

\section{Related Work}
There is so much related work on GANs \citep{GAN} that it is impossible to do it justice 
(see \citet{openquestions,compare_gan} for different overviews of the field),
but here we sketch out a few different threads.
There is a several-year-long thread of work on scaling GANs up to do conditional image synthesis on the ImageNet-2012 
data set beginning with \citet{ACGAN}, extending through \citet{SNGAN, SAGAN, BigGAN, ylg} and most recently culminating
in \citet{LOGAN} and \citet{CRGAN}, which presently represent the state-of-the-art models at this task 
(\citet{LOGAN} uses a larger model size than \citet{CRGAN} and correspondingly report better scores).

\citet{zhou2019don} try to make the discriminator robust to adversarial attacks to the generated images. Our zCR is different in two aspects: zCR enforces the robustness of the compound function $D(G(*))$ to make $D(G(z))$ and $D(G(z + \Delta z))$ consistent, while \citet{zhou2019don} only encourage the robustness in the generated image space as they regularize between $D(G(z))$ and $D(G(z) + v)$ where $v$ is a fast normalized gradient attack vector; Instead of only regularizing $D$, zCR also regularizes $G$ to make $G(z)$ and $G(z + \Delta z)$ different to avoid mode collapse. 

% There is a separate thread of more `graphics-focused' work on GANs that tends not to use the same benchmarks 
% and is hard to directly compare with~\citep{PGGAN, stylegan, styleganv2}, but nevertheless produces
% interesting and impressive results.
% Finally, as GANs are known to be hard to train for a variety of reasons, 
% there is a substantial amount of work 
% \citep{unrolled, wgan, gulrajani2017improved, skillrating, smallgan} dedicated to fixing these 
% issues, understanding them better, or more accurately measuring the quality of GAN outputs.

Most related work on consistency regularization is from the semi-supervised learning literature,
and focuses on regularizing model predictions to be invariant to small perturbations 
\citep{bachman2014learning, CONSISTENCY, laine2016temporal, miyato2018virtual, UDACONSISTENCY} for the purpose
of learning from limited labeled data.
\citet{improvingimproving, CRGAN} apply related ideas to training GAN models and observe initial gains,
which motivates this work.

There are also several concurrent work related to this paper, indicating an emerging direction of GAN training with augmentations. 
\citet{zhao2020differentiable} and \citet{karras2020training} research on how to train GANs with limited data;
while \citet{zhao2020image} mainly focus on thoroughly investigating the effectiveness of different types of augmentations.

\section{Conclusion}

Extending the recent success of consistency regularization in GANs~\citep{improvingimproving,CRGAN}, 
we present two novel improvements: 
Balanced Consistency Regularization, in which generator samples are also augmented along with 
training data, and Latent Consistency Regularization, in which 
draws from the prior are perturbed, 
and the sensitivity to those perturbations is discouraged and 
encouraged for the discriminator and the generator, respectively.

In addition to fixing a new issue we observe with the vanilla Consistency Regularization (augmentation 
artifacts in samples), our techniques yield the best known FID numbers for both unconditional
and conditional image synthesis on the CIFAR-10 data set.
They also achieve the best FID numbers (with the fixed number of parameters used in the 
original BigGAN \citep{BigGAN} model) for conditional image synthesis on ImageNet.

These techniques are simple to implement, not particularly computationally burdensome, and relatively
insensitive to hyper-parameters. 
We hope they become a standard part of the GAN training toolkit, and that their use allows more interesting usage of GANs to many sorts of applications.

% \section*{Software and Data}

% If a paper is accepted, we strongly encourage the publication of software and data with the
% camera-ready version of the paper whenever appropriate. This can be
% done by including a URL in the camera-ready copy. However, \textbf{do not}
% include URLs that reveal your institution or identity in your
% submission for review. Instead, provide an anonymous URL or upload
% the material as ``Supplementary Material'' into the CMT reviewing
% system. Note that reviewers are not required to look at this material
% when writing their review.

\section*{Acknowledgements}
We would like to thank Pouya Pezeshkpour and Colin Raffel for helpful discussions.
This work is funded in part by the National Science Foundation grant IIS-1756023.

% \textbf{Do not} include acknowledgements in the initial version of
% the paper submitted for blind review.

% If a paper is accepted, the final camera-ready version can (and
% probably should) include acknowledgements. In this case, please
% place such acknowledgements in an unnumbered section at the
% end of the paper. Typically, this will include thanks to reviewers
% who gave useful comments, to colleagues who contributed to the ideas,
% and to funding agencies and corporate sponsors that provided financial

\bibliography{example_paper}

\clearpage
\appendix

\section{GAN Losses}
\label{sec:background}

A Generative Adversarial Network (GAN)~\citep{GAN} is composed of a Generator model, $G$,
and a Discriminator model, $D$, which are parameterized by
deep neural networks.
The generator is trained to take a latent vector $z \sim p(z)$ from a prior distribution and generate target samples $G(z)$. 
The discriminator is trained to distinguish
samples from the target distribution $p_\text{real}(x)$ and samples $G(z)$, 
which encourages generator to reduce the discrepancy between the target distribution and $G(z)$. 
Both models have respective losses defined as:
\begin{equation*}
\begin{split}
L_D&=-\mathbb{E}_{x \sim p_\text{data}}\left[\log D(x)\right] - \mathbb{E}_{z \sim p_z
}\left[\log (1 -  D(G(z)))\right], \\
L_G&=- \mathbb{E}_{z \sim p_z}\left[\log D(G(z))\right]. 
\end{split} \label{eq:ns_gan}
\end{equation*}
% where $p(z)$ is usually a standard normal distribution. 
This original formulation~\citep{GAN} is known as the non-saturating (NS) GAN. 
Extensive research has demonstrated that appropriate re-design of $L_D$ plays an important role in training stability and generation quality.
For example, the hinge loss on the discriminator \citep{lim2017, Tran2017} is defined as:
\begin{equation*}
\begin{split}
L_D&=-\mathbb{E}_{x \sim p_\text{data}}\left[\min(0, -1+D(x))\right] \\
& - \mathbb{E}_{z \sim p_z}\left[\min(0, -1-D(G(z)))\right], \\
L_G&=- \mathbb{E}_{z \sim p_z}\left[D(G(z))\right].
\end{split} \label{eq:hinge_gan}
\end{equation*}
The Wassertein GAN (WGAN) \citep{wgan} is another successful reformulation which measures 1-Lipschitz constrained Wasserstein distance \citep{villani2008optimal} between the target distribution and the generated distribution in the discriminator output space.
The objectives of WGAN can be defined as:
\begin{equation*}
\begin{split}
L_D&=-\mathbb{E}_{x \sim p_\text{data}}\left[D(x)\right] + \mathbb{E}_{z \sim p_z}\left[D(G(z))\right], \\
L_G&=- \mathbb{E}_{z \sim p_z}\left[D(G(z))\right].
\end{split} \label{eq:wass_gan}
\end{equation*}
% WGAN also proposes weight clipping to let $D$ satisfy 1-Lipschitz constraint. 
Follow-up work improves WGAN in multiple ways~\citep{gulrajani2017improved,improvingimproving}.
For instance, \citet{SNGAN} propose spectral normalization to stabilize the training, which is widely used~\citep{SAGAN,BigGAN,chen_dataset,chen_temporal}
and has become the de-facto weight normalization technique for GANs.

\section{Evaluation in Inception Score}

Inception Score (IS) is another GAN evaluation metric introduced by~\citet{salimans2016improved}. 
Here, we compare the Inception Score of the unconditional generated samples on CIFAR-10 and CelebA for the experiments in Section~\ref{sec:unconditional}. 
As shown in Table~\ref{tab:is}, our Improved Consistency Regularization achieves the best IS result with both SNDCGAN and ResNet architectures.

\newpage

\begin{figure}[hb]
  \centering
  \subfloat[SNDCGAN on CIFAR-10 with hinge loss.]
  {\includegraphics[width=\columnwidth]{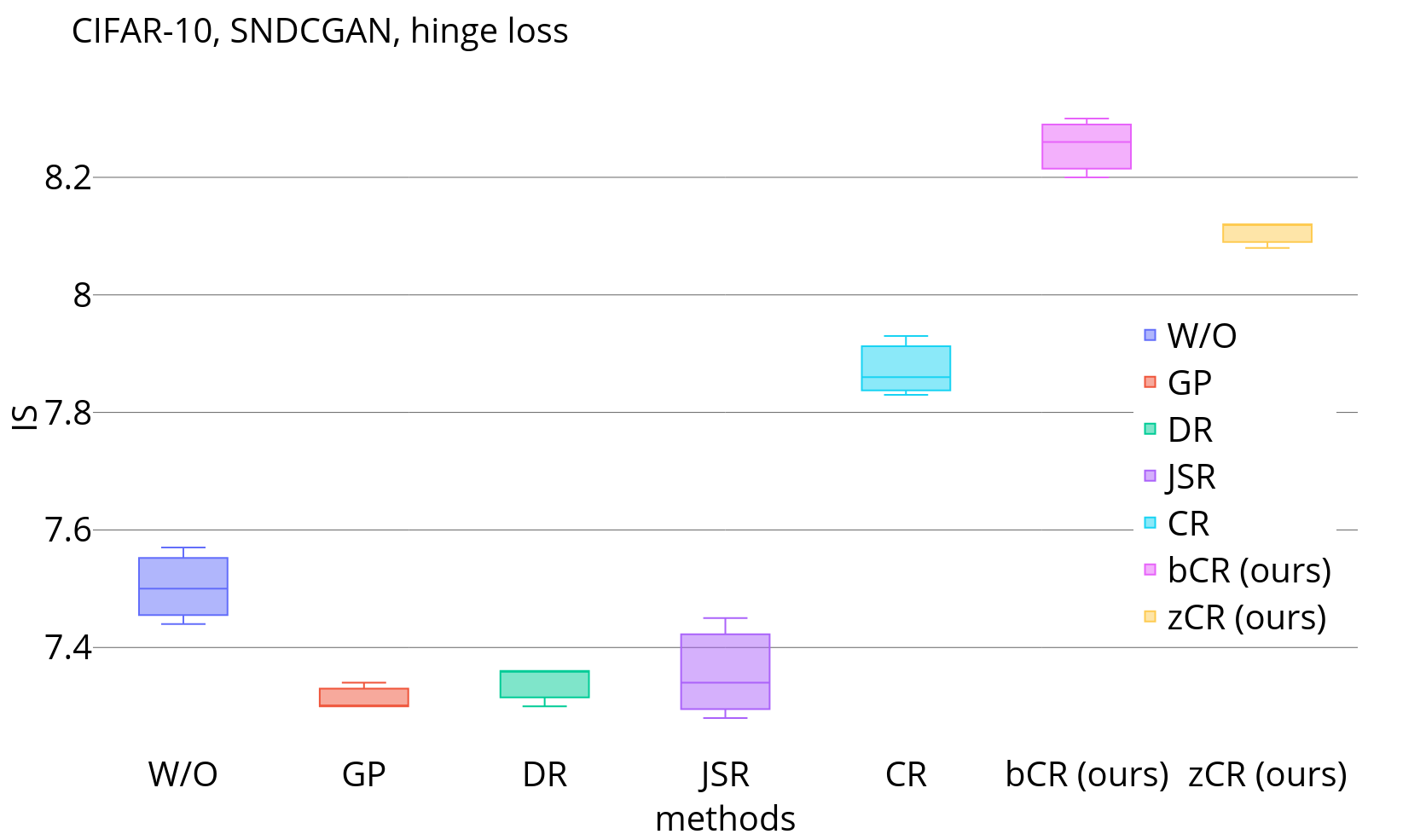}}  \hfill
  \subfloat[ResNet on CIFAR-10 with non-saturating loss.]
  {\includegraphics[width=\columnwidth]{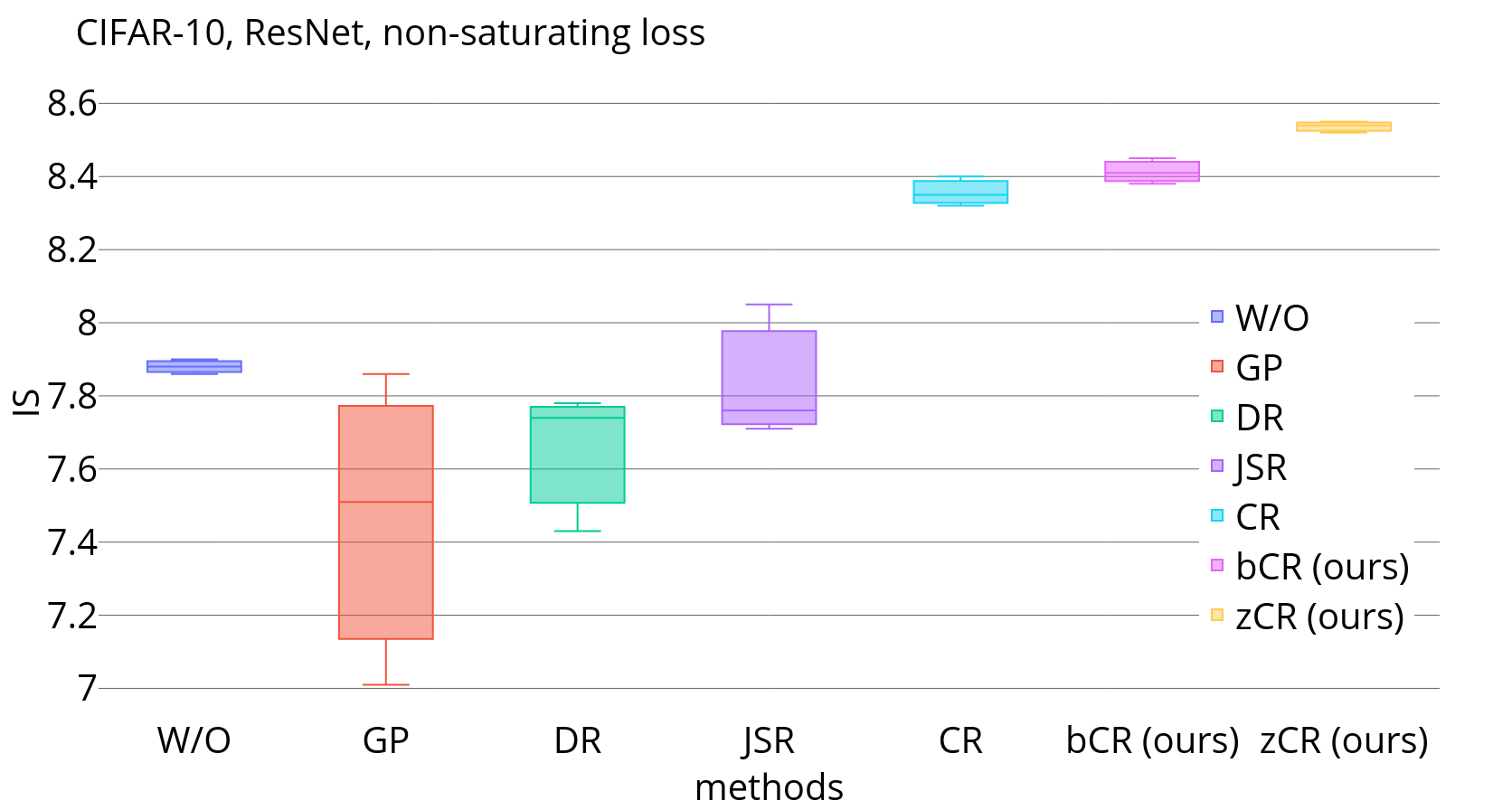}}  \hfill  
  \subfloat[SNDCGAN on CelebA with non-saturating loss.]
  {\includegraphics[width=\columnwidth]{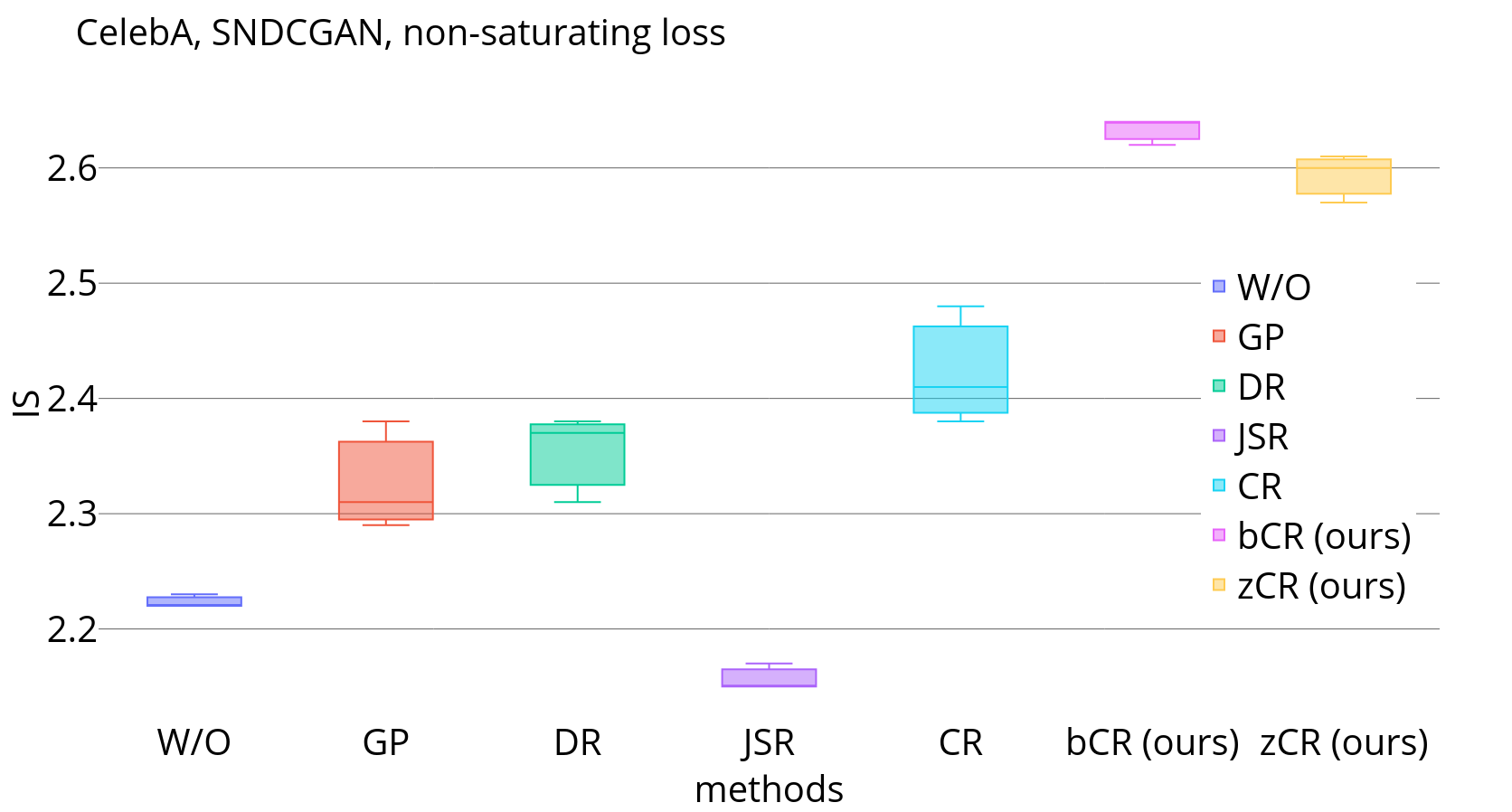}} 
\caption{
Inception Scores.
}
\end{figure}

\begin{table}[htb]
\caption{Best Inception Scores of unconditional image generation.}
\label{tab:is}
\begin{center}
\begin{tabular}{lccc}
\toprule
        & CIFAR-10  & CIFAR-10  & CelebA \\
Methods & (SNDCGAN) & (ResNet)  & (SNDCGAN) \\
\midrule
W/O     & 7.54 & 8.20 & 2.23 \\
GP      & 7.54 & 8.04 & 2.38 \\
DR      & 7.54 & 8.09 & 2.38 \\
JSR     & 7.52 & 8.03 & 2.17 \\
CR      & 7.93 & 8.40 & 2.48 \\
\midrule
ICR (ours)    & \textbf{8.14} & \textbf{8.55} & \textbf{2.64} \\
\bottomrule
\end{tabular}
\end{center}
\end{table}

\newpage
\section{Hyper-Parameters' Effect on Performance}

In Section~\ref{sec:discussion}, we carry out extensive experiments analyzing how the hyper-parameters affect performance for Balanced Consistency Regularization and Latent Consistency Regularization respectively. 
Due to space limit, we present the visualization of results here in the appendix.

\begin{figure}[hb!]
\begin{center}
\centerline{\includegraphics[width=0.8\columnwidth]{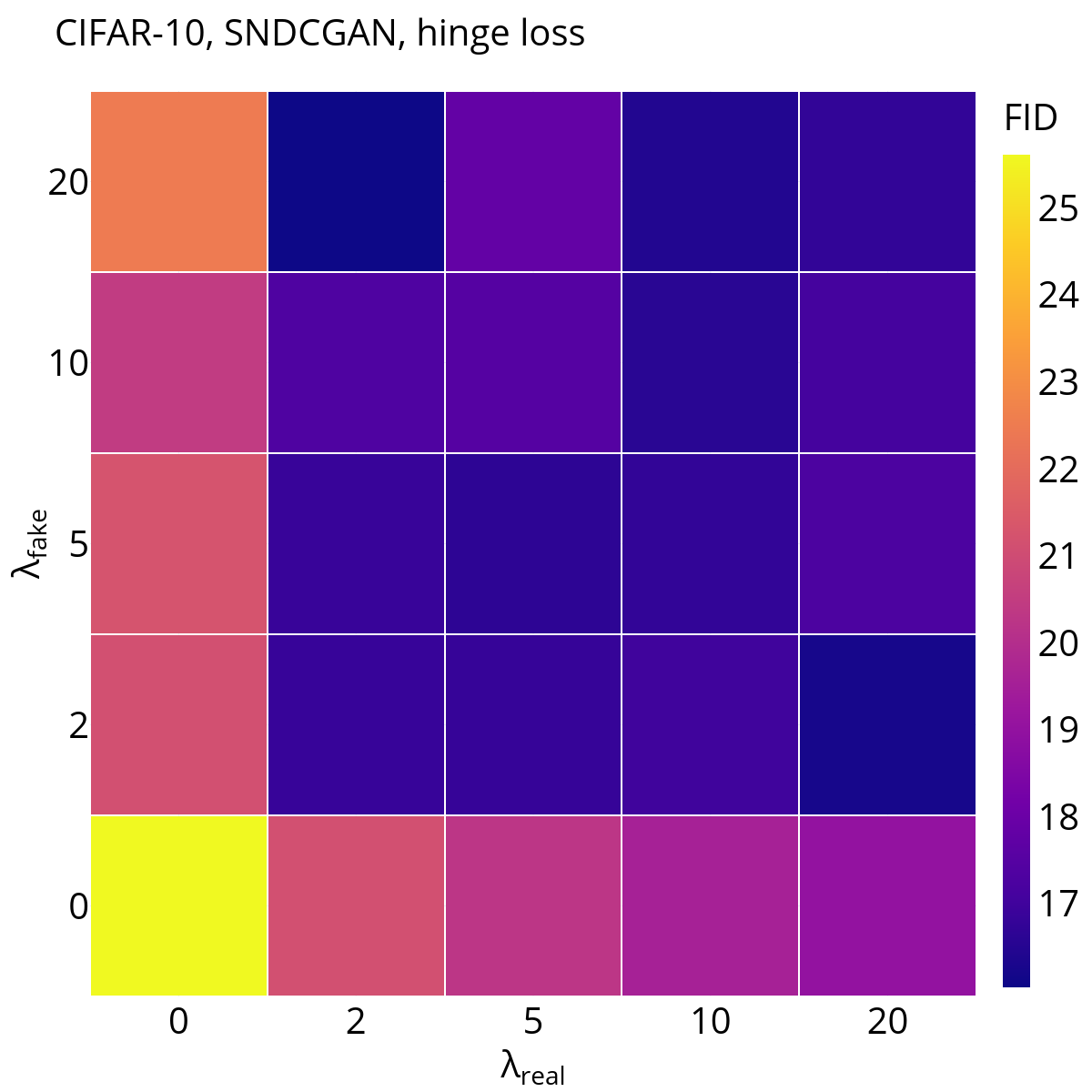}}
\caption{
Analysis of the effects of the $\lambda_\text{real}$ and $\lambda_\text{fake}$
hyper-parameters for Balanced Consistency Regularization.
We train DCGAN on CIFAR-10 with hinge loss, 
for many different values of $\lambda_\text{fake}, \lambda_\text{real}$.
The results show that Balanced Consistency Regularization essentially always outperforms
vanilla consistency regularization.
Generally speaking, Balanced Consistency Regularization performs best with
$\lambda_\text{fake}$ and $\lambda_\text{real}$ of similar magnitudes.
}
\label{fig:real_fake_ratio}
\end{center}
\end{figure}

\section{Additional Quantitative Results}

In Section~\ref{sec:discussion:artifact}, we examine artifacts resulted from `Vanilla' Consistency Regularization in \citet{CRGAN}, and demonstrate that our Balanced Consistency Regularization (bCR) can alleviate the issues with illustrations.
Here in the appendix, we report quantitative evaluations in terms of FID corresponding to the experiments in Figure~\ref{fig:artifact_analysis}.

\begin{table}[tb]
\caption{FID of CIFAR-10 augmented with cutout corresponding to Figure~\ref{fig:artifact_analysis}.}
\label{tab:appx_cutout}
\begin{center}
\begin{tabular}{lcc}
\toprule
FID      & CR & Our bCR \\
\midrule
    $8 \times 8$ cutout & $20.12 \pm 0.29$ & $16.65 \pm 0.22$ \\
    $16 \times 16$ cutout & $29.24 \pm 0.72$ & $17.08 \pm 0.61$ \\
    $32 \times 32$ cutout & $38.03 \pm 1.01$ & $17.94 \pm 0.48$ \\
\bottomrule
\end{tabular}
\end{center}
\end{table}

Moreover, we further carry out experiments with a different augmentation: color jittering, the artifacts resulted from which can also be easy to perceive.
For augmentation with color jittering, we add random noise constrained by a certain strength to image color channels.
For example, with the color jittering strength of $0.3$, random noise between $(0, 0.3)$ is added onto normalized images for augmentation.
Table~\ref{tab:appx_color} shows we have similar results when augmenting images with color jittering comparing CR with bCR, which support our findings and improvements over baselines.

\begin{table}[hb]
\caption{FID of CIFAR-10 augmented with color jittering. }
\label{tab:appx_color}
\begin{center}
\begin{tabular}{lcc}
\toprule
FID      & CR & Our bCR \\
\midrule
    $0.1$ strength & $18.22 \pm 0.27$ & $16.24 \pm 0.25$ \\
    $0.2$ strength & $21.07 \pm 0.33$ & $17.56 \pm 0.41$ \\
    $0.3$ strength & $27.16 \pm 0.58$ & $18.14 \pm 0.68$ \\
\bottomrule
\end{tabular}
\end{center}
\end{table}

% \newpage
\section{Qualitative Examples}

We randomly sample from our ICR-BigGAN model on ImageNet (FID=5.38, Secition~\ref{sec:conditional}) as qualitative examples for different class labels.
We have obtained permission from authors of CR-GAN~\citep{CRGAN} to directly use the visualization of random samples from their CR-BigGAN model (FID=6.66) for comparison.
In the following figures, the left column shows random samples from our ICR-BigGAN, while the right column presents those from baseline CR-BigGAN.

%%%%%%%%%%
\clearpage
%%%%%%%%%%

\begin{figure*}[hbtp]
  \centering
  \subfloat[Monarch Butterfly (our ICR vs baseline CR)]
  {\includegraphics[width=0.78\columnwidth]{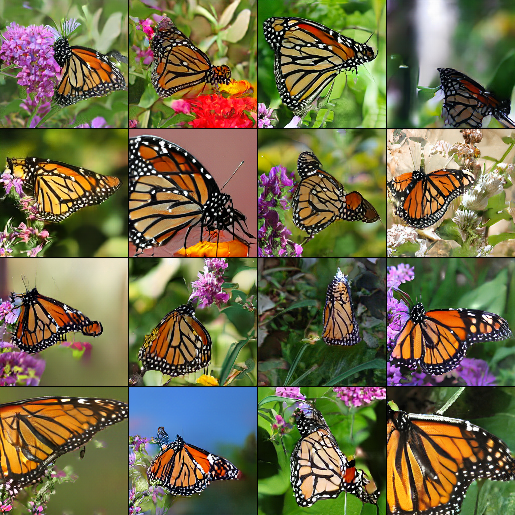} \hspace{1 cm}
   \includegraphics[width=0.78\columnwidth]{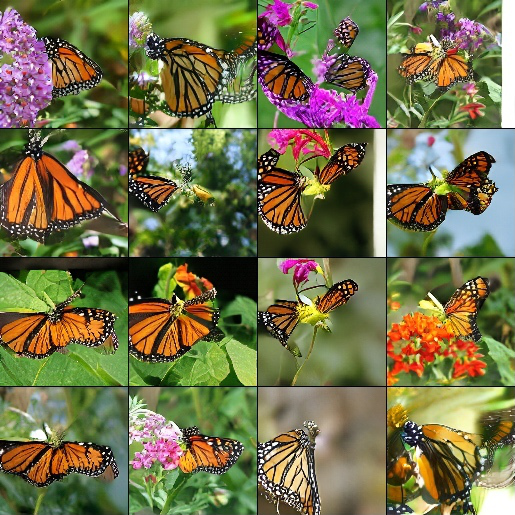}} \hfill
\subfloat[Cock (our ICR vs baseline CR)]
  {\includegraphics[width=0.78\columnwidth]{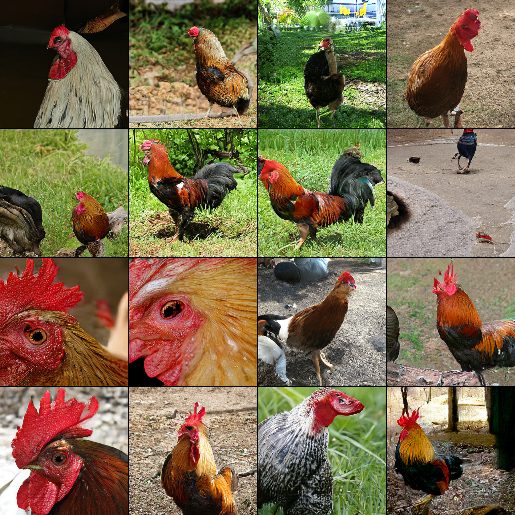} \hspace{1 cm}
   \includegraphics[width=0.78\columnwidth]{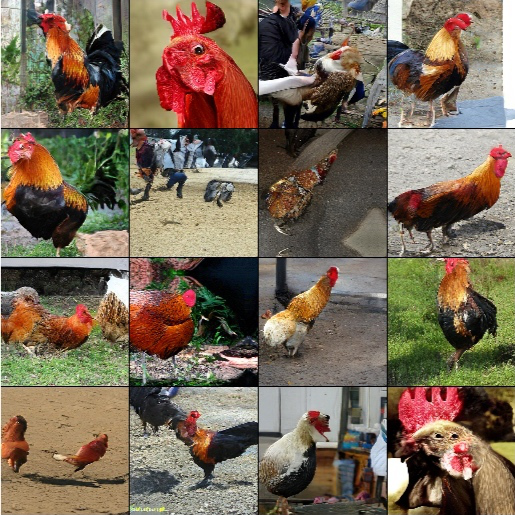}} \hfill
  \subfloat[Blenheim Spaniel (our ICR vs baseline CR)]
  {\includegraphics[width=0.78\columnwidth]{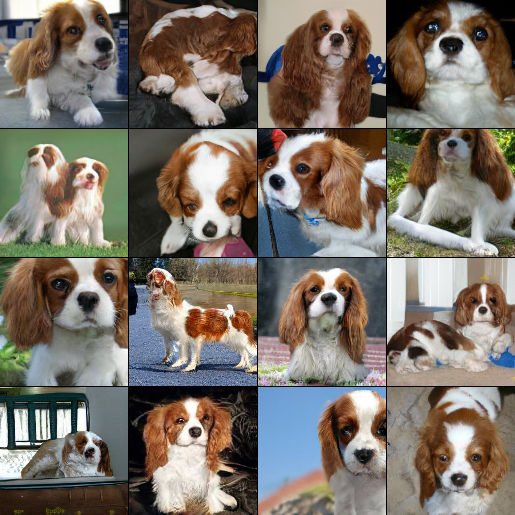} \hspace{1 cm}
   \includegraphics[width=0.78\columnwidth]{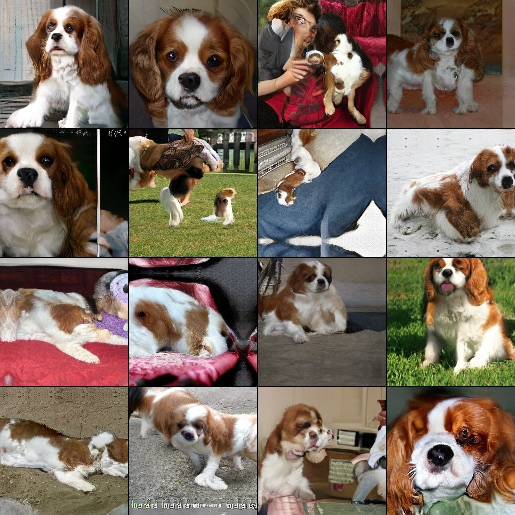}}   
\caption{
Random ImageNet samples from our ICR-BigGAN (
% Section~\ref{sec:conditional}, 
FID 5.38) vs CR-BigGAN (\citet{CRGAN}, FID 6.66).
}
\end{figure*}

\begin{figure*}[hbtp]
  \centering
  \subfloat[Cheeseburger (our ICR vs baseline CR)]
  {\includegraphics[width=0.78\columnwidth]{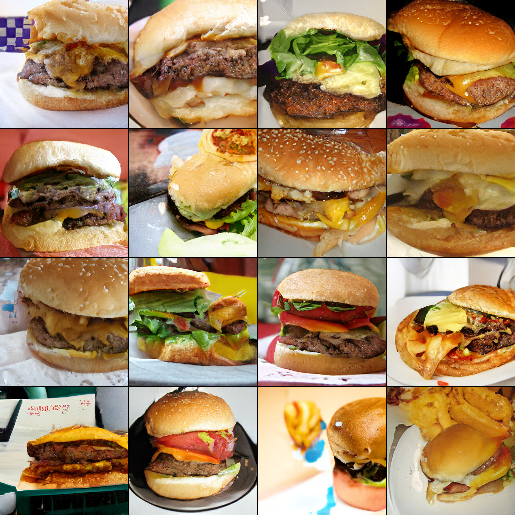} \hspace{1 cm}
   \includegraphics[width=0.78\columnwidth]{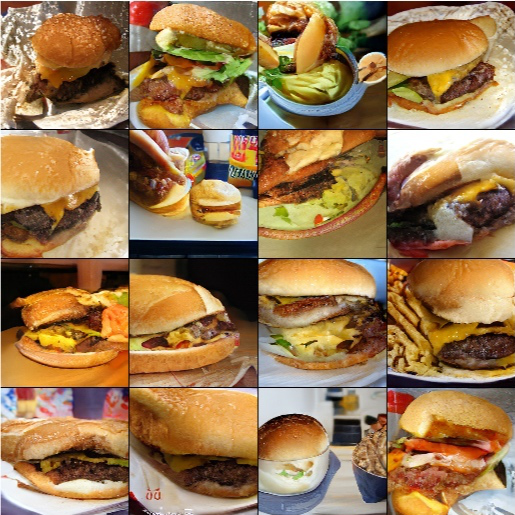}} \hfill
  \subfloat[Ambulance (our ICR vs baseline CR)]
  {\includegraphics[width=0.78\columnwidth]{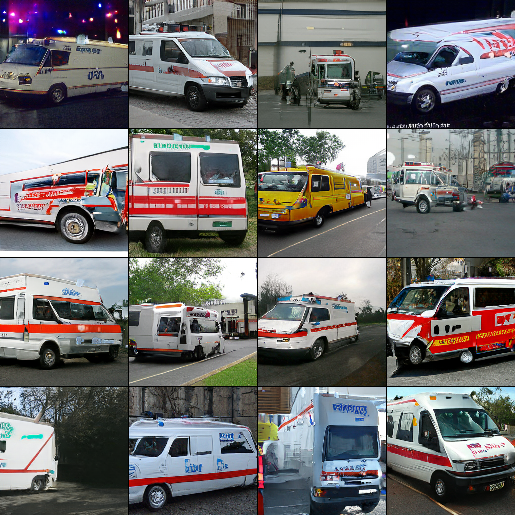} \hspace{1 cm}
   \includegraphics[width=0.78\columnwidth]{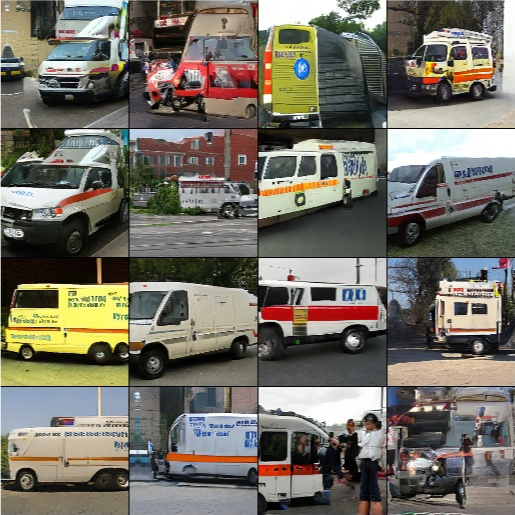}} \hfill
  \subfloat[Beer Bottle (our ICR vs baseline CR)]
  {\includegraphics[width=0.78\columnwidth]{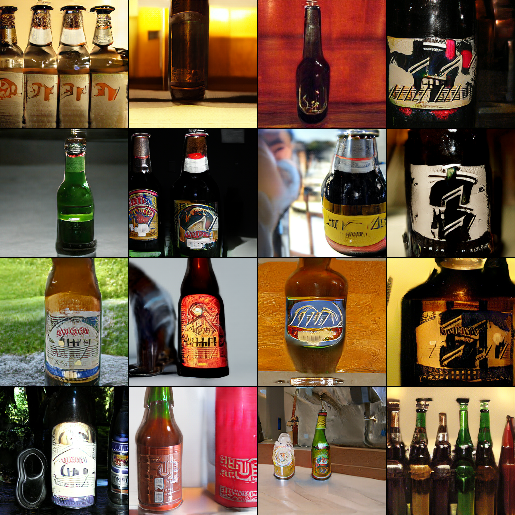} \hspace{1 cm}
   \includegraphics[width=0.78\columnwidth]{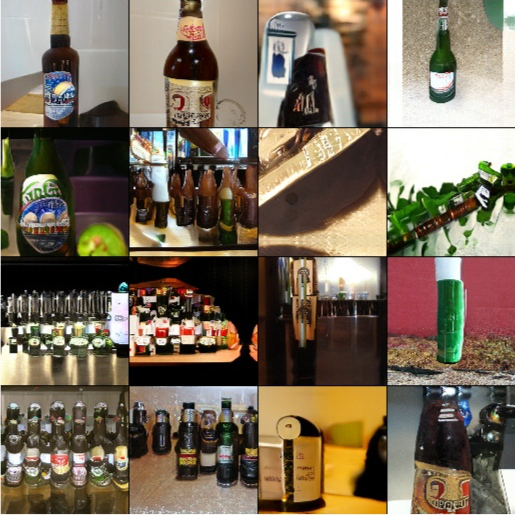}}   
\caption{
Random ImageNet samples from our ICR-BigGAN (
% Section~\ref{sec:conditional}, 
FID 5.38) vs CR-BigGAN (\citet{CRGAN}, FID 6.66).
}
\end{figure*}

\begin{figure*}[hbtp]
  \centering
  \subfloat[Husky]
  {\includegraphics[width=2\columnwidth]{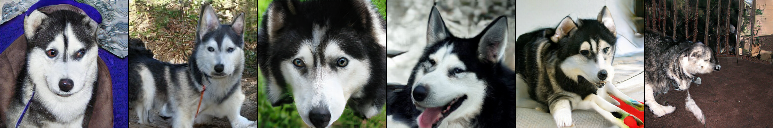}} \hfill
  \subfloat[Car Mirror]
  {\includegraphics[width=2\columnwidth]{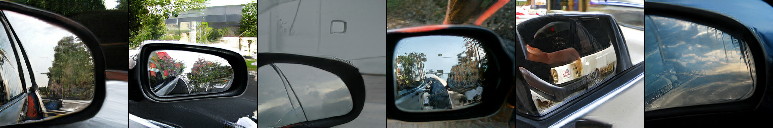}} \hfill
  \subfloat[Barn]
  {\includegraphics[width=2\columnwidth]{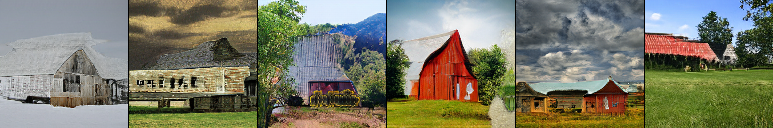}} \hfill
  \subfloat[Forklift]
  {\includegraphics[width=2\columnwidth]{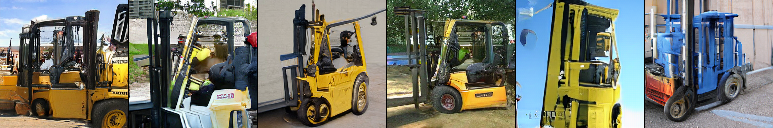}} \hfill
  \subfloat[Daisy]
  {\includegraphics[width=2\columnwidth]{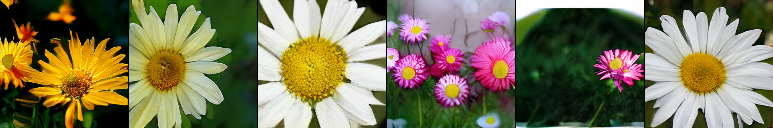}} \hfill
  \subfloat[Tabby Cat]
  {\includegraphics[width=2\columnwidth]{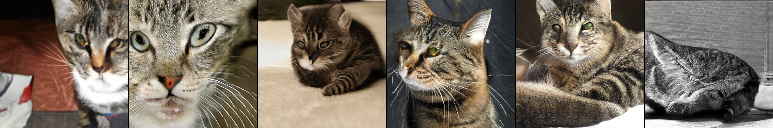}}    
\caption{
More random samples from our ICR-BigGAN (Section~\ref{sec:conditional}) trained on ImageNet.
}
\end{figure*}

\end{document}